# Performing Hybrid Recommendation in Intermodal Transportation – the FTMarket System's Recommendation Module

**Alexis Lazanas**

**Industrial Management and Information Systems Lab, University of Patras**
**Rion Patras, 26500, Greece**
*alexlas@mech.upatras.gr*

**Abstract**
Diverse recommendation techniques have been already proposed and encapsulated into several e-business applications, aiming to perform a more accurate evaluation of the existing information and accordingly augment the assistance provided to the users involved. This paper reports on the development and integration of a recommendation module in an agent-based transportation transactions management system. The module is built according to a novel hybrid recommendation technique, which combines the advantages of collaborative filtering and knowledge-based approaches. The proposed technique and supporting module assist customers in considering in detail alternative transportation transactions that satisfy their requests, as well as in evaluating completed transactions. The related services are invoked through a software agent that constructs the appropriate knowledge rules and performs a synthesis of the recommendation policy.

## 1. Introduction

Transportation management involves diverse decision making issues, which are basically related to the appropriate route and carrier selection. Such issues mainly raise due to the variety of the customer's preferences (e.g. cost limitations, loading preferences, delivery dates) and the carrier's service resources (e.g. transportation media, available itineraries, capacity). The matching between the above preferences and offered services cannot be easily handled manually, as in most cases a plethora of alternative options exist, while time and money limitations are ubiquitous. Generally speaking, transportation transactions management requires quick and cost-effective solutions to the customers' demands for both distribution and shipping operations. In cases where many alternatives exist, there is an urgent need for providing recommendations. The customer should be assisted in order to properly evaluate the proposed alternatives and make his/her final decision.

Recommendation systems have been described as systems that produce individualized recommendations or have the effect of guiding the user in a personalized way, in environments where the amount of on-line information vastly outstrips any individual's capability to survey it [2]. Generally speaking, such systems represent the users' preferences for the purpose of submitting suggestions for purchasing or evaluating elements. Fundamental applications can be found in the fields of electronic commerce and information retrieval, where they provide suggestions that effectively direct the users to the elements that satisfy better their necessities and preferences [21].

This paper reports on the development of an innovative recommendation module that provides valuable assistance to the users of a transportation transactions management system, namely FTMarket (Freight Transportation Market). FTMarket is fully implemented and handles various types of transportation transactions [14, 10]. It exploits a series of dedicated software agents that represent and act for any type of user involved in a transportation scenario (such as customers who look for efficient ways to ship their products and transport companies that may - fully or partially - carry out such requests), while they cooperate and get the related information in real-time mode [24]. Our overall approach is based on flexible models that achieve efficient communication among all parties involved, coordinate the overall process, construct possible alternative solutions and perform the required decision-making [10, 12]. In addition, FTMarket is able to handle the complexity that is inherent in such environments [6], which concerns freighting and fleet scheduling processes, as well as "modular transportation solutions"[1]. FTMarket provides

---
[1] To further explain this concept, consider the case where a customer wants to convey some goods from place A to place B, while there is no transport company acting directly between these two places. Supposing that two available carriers X and Y have some scheduled itineraries from A to C and from C to B, respectively, it is obvious that a possible solution

the customer with a set of alternative solutions for each requested transaction. These solutions are constructed through the use of a specially developed algorithm for retrieving optimal and sub-optimal solutions. Moreover, through a dedicated recommender agent [9, 22], which builds on Web Services concepts [26], the system assists the customer further towards making the appropriate decisions.

The remainder of this paper is structured as follows: Section 2 reports on background issues from the area of recommender systems, paying particular attention to recommendation approaches. Section 3 describes the basic aspects of our approach, which concern the selection of transportation plans and the evaluation of alternative solutions. Section 4 focuses on issues raised during the integration of the recommendation module, the formulation of the recommendation policy, and the exploitation of software agents and Web Services technologies. Finally, Section 5 concludes the paper and highlights future work directions.

## 2. Related Work

The most widely adopted recommendation techniques are Collaborative Filtering (CF) and Knowledge Based Recommendation (KBR), each one possessing its own strengths and weaknesses. Collaborative Filtering (CF) [17, 18] is the most commonly used recommendation technique to date. The basic idea of CF-based algorithms is to provide item recommendations or predictions, based on the opinion of other like-minded users. In a typical CF scenario, there is a list of m users $U = \{u_1, u_2, ..., u_m\}$ and a list of $n$ items $I = \{i_1, i_2, ..., i_n\}$. Each user $u_i$ is associated with a list of items $Iu_i$, for which the user has expressed his/her opinion. Opinions can be explicitly given by the user as a rating score (within a certain numerical scale), or implicitly derived from transaction records (by analyzing timing logs, mining web hyperlinks and so on). For a particular user $u_a$, the task of a collaborative filtering algorithm is to find an *item likeness* that can be of two forms:

- *Prediction:* this is a numerical value, $P_i$, expressing the predicted likeness of item $i$ ($i$ does not belong to $Iu_a$) for the user. The predicted value is within the same scale (e.g. from 1 to 5) as the opinion values provided by $u_a$ [19].

- *Recommendation:* this is a list of $N$ items $I_r$ ($I_r$ is a subset of $I$) that the user will like most (the recommended list must contain items not already selected by the user). This outcome of CF algorithms is also known as *Top-N* recommendation [20].

On the other hand, KBR attempts to suggest objects based on inferences about a user's needs and preferences. In some sense, all recommendation techniques could be described as doing some kind of inference. Knowledge-based approaches are distinguished in that they utilize functional knowledge; in other words, they have knowledge about how a particular item meets a particular user need and can therefore reason about the relationship between a need and a possible recommendation. The user profile can be any knowledge structure that supports this inference. In the simplest case, as in *Google*, it may simply be the query that the user has formulated. The *Entrée* system and several other recent systems [23], employ techniques from case-based reasoning for knowledge-based recommendations.

The knowledge used by a knowledge-based recommender system can take many forms. *Google* uses information about the links between web pages to infer popularity and authoritative value [1]. *Entrée* uses knowledge of cuisines to infer similarity between restaurants. Utility-based approaches calculate a utility value for objects to be recommended; in principle, such calculations could be based on functional knowledge. However, existing systems do not use such inference mechanisms, thus requiring users to do their own mapping between their needs and the features of products, either in the form of preference functions for each feature, as in the case of *Tête-à-Tête*, or answers to a detailed questionnaire, as in the case of *PersonaLogic* [2]. Knowledge-based recommender systems are prone to the drawback of all knowledge-based systems: the need for knowledge acquisition. More specifically, there are three types of knowledge that are involved in such systems:

- *Catalog knowledge:* Knowledge about the objects being recommended and their features. For example, the system should know that "Gasoline" is a type of "Fuel".
- *Functional knowledge:* The system must be able to match the user's needs with the object that might satisfy those needs. For example, a recommendation module should know that the transportation of toxics require a higher safety level.
- *User knowledge:* To provide good recommendations, the system must have some knowledge about the user.

---

to the above customer's request is to involve both X and Y and fragment the intended overall itinerary to the related sub-routes. It is also noted that these carriers may be associated with diverse transportation means, such as trains, trucks, ships and airplanes.

This might take the form of general demographic information or specific information about the need for which a recommendation is sought.

Of these knowledge types, the last one is the most challenging, as it is an instance of the general user-modelling problem [25]. Despite this drawback, knowledge-based recommendation has some beneficial characteristics. First of all, it is appropriate for casual exploration, because it demands less from the user (compared to the utility-based recommendation). Moreover, it does not involve a start-up period during which its suggestions are of low quality. On the other hand, a knowledge-based recommender cannot "discover" user niches, the way collaborative systems can. However, it can make recommendations as wide-ranging as its knowledge base allows.

Alternative techniques have been proposed in the literature in order to handle the above issues [11]. Having thoroughly considered their pros and cons, our approach follows a hybrid recommendation technique. Generally speaking, CF and KBR techniques can be combined in hybrid recommendation systems in order to improve their performance. Most commonly, CF is combined with some other technique in an attempt to minimize or avoid the ramp-up problem [3].

## 3. The Proposed System

3.1 Transportation plans and evaluation of alternative solutions

The recommendation procedure adopted in our approach is highly associated with the selection (by the user) of the appropriate transportation plan. A transportation plan typically defines the user preferences for the upcoming transactions. The five alternative plans offered are:

- Express
- Economic
- Safe
- Dependable
- User Defined

It can be easily observed that each of the first four plans declares a specific tension in the recommendation strategy to be followed by the system, in that it either minimizes the overall duration or cost (first two plans), or it retains a high level of safety or dependability (third and fourth plans) of the suggested itineraries. The last choice offers the possibility for a user-customized plan definition. Such a plan may combine parameters from all the above four plans. The selection of one of these plans will influence the recommendation process of our approach for the particular user.

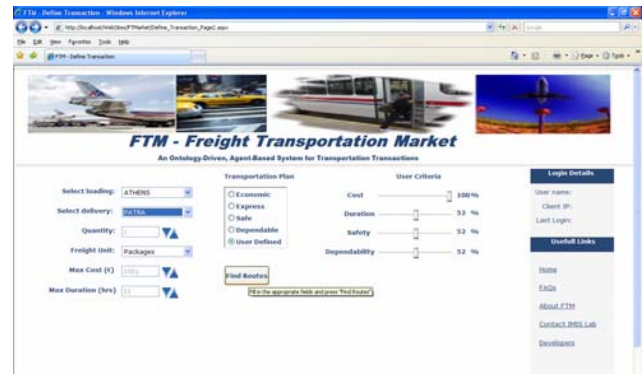

Figure 1: Transaction's request interface

As shown in Figure 1, which depicts the system's interface for handling a user's request, the user provides input about the loading and delivery terminals, the quantity to be transported, expresses his/her preferences concerning maximum cost and duration of the transaction, and selects the desired transportation plan. By selecting the "user-defined" plan, a new window appears, allowing the user to adjust the criteria (cost, duration, safety, dependability) of his/her transportation request.

**Table 1:** Selection criteria for the alternative transportation plans
*(safety and dependability take values from the set {very low, low, average, high, very high}).*

| *Plan* | Cost | Duration | Safety | Dependability |
|---|---|---|---|---|
| *Express* | Any | Min | Any | Any |
| *Economic* | Min | Any | Any | Any |
| *Safe* | Any | Any | >Average | ≥Low |
| *Dependable* | Any | Any | ≥ Low | > Average |
| *Hybrid* | User Defined | User Defined | User Defined | User Defined |

During the construction of the available transportation solutions, our approach excludes solutions that do not comply with the customer's requirements. More specifically, a set of predefined rules is employed to exclude the alternative solutions that do not correspond to the specific freight transportation's requirements and customer preferences. Table 1 summarizes the constraints to be met for each transportation plan (for the *"User Defined"* plan, this process takes into account the constraints set by the user). In all cases, solutions that do not satisfy these constraints are discarded.

## 3.2 A Methodology for the Selection of Alternative Route Paths

In our former work [10, 27], we have presented an algorithm for constructing optimal (direct or modular) solutions for a requested transportation transaction. This algorithm was taking into account the cost and duration of each sub-route, as well as the cost and duration upper bounds (as they had been set by the user). If no optimal solution could be constructed, the algorithm terminated without providing any solutions. To better handle such cases, our approach uses an elaborated version of Dijkstra's shortest path algorithm [4] to construct sub-optimal solutions. Even if such solutions cannot be characterized as optimal, they represent acceptable alternatives for a specific transportation request.

As it can be retrieved from the related literature [4], shortest path algorithms use a bidirectional, single-weighted graph to represent a connected set of vertices ($V_i$) through a number of arcs $A_{ij}$ (from $V_i$ to $V_j$). Our algorithm takes into consideration each $A_{ij}$ and its correspondent weight ($W_{ij}$) in order to produce a route path from a starting point ($S$) to an ending point ($E$) that minimizes the total weight ($W_{SE}$). The complexity of our approach consists in the presence of a pair of variables that affect each arc's weight, namely the cost and the duration. Due to the fact that there exist two weights for each arc (cost and duration), we confronted the problem of unifying these weights into a single one, in order to proceed with the ranking of the solutions. As shown in Figure 2, each arc's $A_{ij}$ weight ($W_{ij}$) consists of a cost weight ($W_{cost-ij}$) and a duration weight ($W_{duration-ij}$). It is obvious that:

$$W_{ij} = W_{cost-ij} + W_{duration-ij} \quad (1)$$

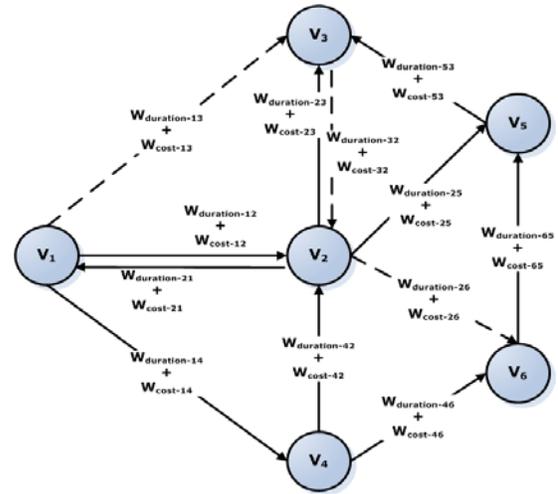

Figure 2: A hypothetical 2-weighted graph.

Having defined the total weight for each arc ($A_{ij}$), we encountered the problem of adding these two parameters that are measured in different units (Euros and hours, respectively). This problem was confronted by applying a normalization technique that divided both the $cost_{ij}$ and $duration_{ij}$ of an arc with its correspondent maximum cost and duration of the sub-route. It is:

$$W_{duration-ij} = \frac{duration_{ij}}{\max(duration_{ij})} \quad (2)$$

$$W_{cost-ij} = \frac{cost_{ij}}{\max(cost_{ij})} \quad (3)$$

Another issue that came up after the weight normalization procedure concerned the solutions' ranking. To address this problem, our approach provides the user with different solutions by using a pair of weight coefficients (`costCoef` and `durationCoef`) and by calculating solutions corresponding to alternative combinations of the weights of the cost and duration criteria (see Figure 3), according to the formula:

$$W_{ij} = (costCoef * W_{cost-ij}) + (durationCoef * W_{duration-ij}) \quad (4)$$

The cost and duration coefficients take values from the set {0, 0.1, 0,2, ..., 1}. The main idea of this process is to provide the algorithm with alternative weights ($w_{ij}$), each one expressing a different combination of cost and duration parameters. At the beginning of this procedure, we calculate the weight of each sub-route by taking into consideration only the duration parameter (we set the cost coefficient to 0 and the duration coefficient to 1). Then, in a step-wise way, we decrease the duration coefficient by 0.1 (obviously, we increase at the same time the cost coefficient by 0.1). Finally, we calculate the sub-route's weight taking into consideration only the cost parameter (the duration coefficient has become 0).

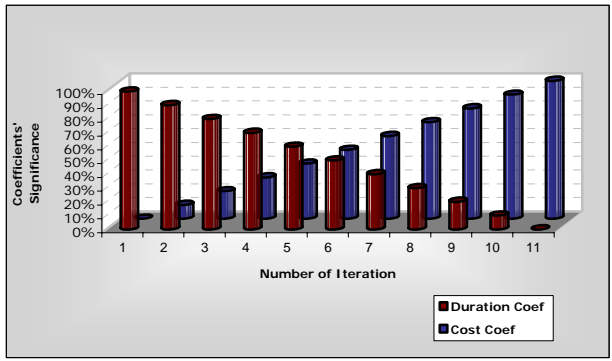

Figure 3: Weight coefficients' variation.

This process is described in pseudo-code as follows:
```
{
costCoef ← 0.0;
durationCoef ← 1.0;
step ← 0.0;
while step ≤ 1.0 calculate
  {
      costCoef ← step;
      durationCoef ← 1-step;
      weight[i][j] ← costCoef*Wcost + 
durationCoef*Wduration;
      perform shortest path algorithm;
      step ← step + 0.1;
  }
}
```

The outcome of the above process is then presented to user. As shown in Figure 4 (which depicts an instance of the related system interface), the optimal routes for a transportation request from Athens to *Patra* have been retrieved (after a related request). The basic characteristics of each route are presented in the main table of the web interface. By selecting the "View Details" option, the user is able to receive an analytical description of the sub-routes contained in each itinerary, as well as their corresponding characteristics. Solutions at this phase are ranked by default according to the cost; in any case, users may request alternative rankings by clicking on the corresponding column header.

Figure 4: Solutions produced by the system.

## 4. Integrating a Recommendation Module

4.1 A Hybrid Recommendation Methodology

The recommendation procedure begins immediately after the abovementioned construction of the alternative solutions. It is a complex process which is carried out in three basic phases, which are:

- the evaluation of the carriers and the transactions data;
- the exploitation of transaction data through a data mining process, and
- the recommendation methodology selection or synthesis.

At the beginning of the process, the system stores all the appropriate data that are submitted by the user and are related with pending or completed transportation transactions. These data are of significant importance and will be further exploited by the data mining process. Moreover, in this phase the user evaluates (i.e. assigns a score to) the carrier(s) involved in a transaction through an appropriate interface.

The second phase of recommendation concerns the data mining process. Data mining is a useful decision support technique, which can be used to find trends and regularities in big volumes of data. At this phase, transactions data are gathered through knowledge construction processes. In our case, the data mining process constructs a model from the recommendation module's database that may produce well defined knowledge rules. This procedure is performed through SQL queries performed on the transactions' tables. After the completion of this process, the constructed knowledge-based rules participate in the production of knowledge-based recommendation data that will be evaluated and synthesized in the last phase of recommendation.

The last phase of recommendation refers to the selection or synthesis of the appropriate recommendation technique. This objective will be reached through the definition of well structured rules that will be applied for each transaction. The *Recommender Agent* of our system takes the initiative to select the most appropriate recommendation technique. For example, for a particular itinerary from point *i* to point *j*, taking into consideration that the customer has selected a certain plan, a rule for the specific itinerary could lead to the recommendation of a carrier that is different than the one suggested by the CF technique, based on the carriers' evaluation process described earlier in this section. The recommendation methodology described above is graphically presented in Figure 5, through a data flow diagram.

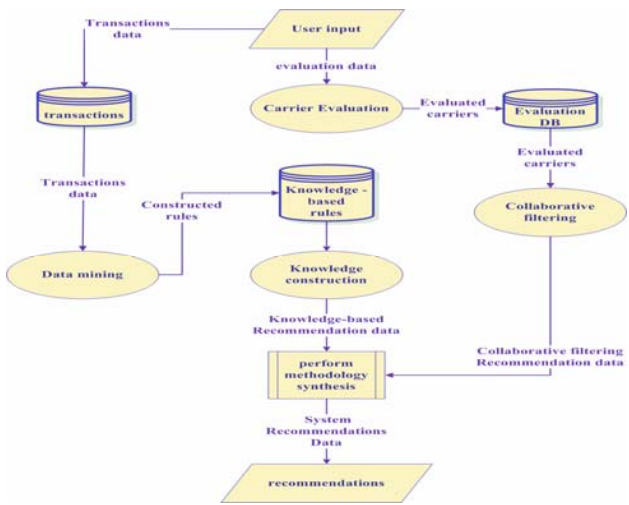

Figure 5: The data flow diagram for the recommendation methodology

Due to the large amount of data the recommendation module takes into account in order to provide knowledge-based recommendations, the database model has been thoroughly considered. The system's database has been designed through the use of SQL Server 2005 Management Console, in order to accomplish further with the customers' needs. Much attention has been paid into the reorganization of data tables' fields, as well as into the representation of the entities' relationships [16]. The database model that participates in the knowledge construction of the recommendation's phase is presented in Table 2.

Table 2: Recommendation Module's Database Model

| Table Name | Description |
|---|---|
| Transactions | Transactions in progress |
| Transaction's Subroutes | Transactions sub-routes in progress |
| Transactions _Rating | Completed Transactions' evaluation |
| Carriers_Rating | Carrier evaluation with completed |
| Users_Reliability | Customers reliability evaluation |
| Temp_Transactions | Proposed transaction itineraries |
| Temp_Transactions_Subroutes | Subroutes of the proposed itineraries |

4.2 Calculation of Recommendation Score

After the ranking phase, the evaluation of each alternative route retrieved is performed. Our system retrieves all possible transportation routes that can be constructed for a given transaction request. These routes are presented to the user through an appropriate designed user interface. The corresponding user interface enables the user to either select one of the proposed routes (in this case, he/she will be asked to complete the transaction), or to be redirected to a user-friendly interface where he/she can receive recommendations for each separate route. The evaluation of a transaction is based on various criteria, such as:

- Cost
- Duration
- Safety
- Reliability
- Average scores of the above carriers' elements.
- Average scores of the sub-routes contained in the transaction
- The number of times that the specific route has been selected by other customers (popularity).
- Number of transloadings

The recommendation procedure is implemented through the evaluation of both the transactions and the transportation companies involved. It is a complex procedure, basically due to the fact that a modular solution may involve two or more carriers. It is obvious that a transaction can receive an overall negative evaluation, while - at the same time - a specific part could have been completed quite satisfactorily. The evaluation of a transaction is based on a set of criteria such as cost, duration, safety, dependability, average score of a carrier, itinerary's popularity and number of transloadings [15]. Taking into consideration all the above issues, we define the calculation formula of the overall score $\left(O_{i,j}^{total}\right)$ of each transaction from point *i* to point *j* (for each sub-route of the itinerary). It is:

$$O_{i,j}^{total} = O_{i,j}^{t} + O_{i,j}^{s} + O_{i,j}^{r} \quad (5)$$

$$O_{i,j}^{final} = \sum_{i,j=1}^{n} \frac{(O_{i,j}^{total} - O_{i,j}^{cost})^2}{f_{S,E}} \quad (6)$$

where $O_{i,j}^t, O_{i,j}^s, O_{i,j}^r$ represent the score of the time, safety and dependability, respectively, for the transportation from point *i* to point *j*. The variable $f_{S,E}$ represents the number of transloadings of each proposed solution and is considered as a negative factor, assuming that a large number of transloadings could evoke damage in the product and increase the transaction's completion time. The number of transloadings is related to the number of sub-routes (*n*) of each itinerary. It is:

$$f_{S,E} = n - 1, n > 1 \quad (7)$$

Each one of the detailed scores is calculated according to the score that has been assigned to the carrier and each sub-route. It is:

$$O_{i,j}^{t} = \frac{\left[avg(C_{i,j}^{t} * ur) + avg(T_t * ur)\right]a}{2} \quad (8)$$

$$O_{i,j}^{s} = \frac{\left[avg(C_{i,j}^{s} * ur) + avg(T_s * ur)\right]b}{2} \quad (9)$$

$$O_{i,j}^{r} = \frac{\left[avg(C_{i,j}^{r} * ur) + avg(T_r * ur)\right]c}{2} \quad (10)$$

where

$C_{i,j}^{t}$ = The carrier's score according to time, for the transportation from point *i* to *j*.

$C_{i,j}^{s}$ = The carrier's score according to safety, for the transportation from point *i* to *j*.

$C_{i,j}^{r}$ = The carrier's score according to dependability, for the transportation from point *i* to *j*.

$T_t$ = The transaction's score according to time.

$T_s$ = The transaction's score according to safety.

$T_r$ = The transaction's score according to dependability.

The expression *avg(x)* refers to the average value of the element *x* in the database, and the variables *a,b,c* are coefficients related with the user's preferences according to time, safety and dependability respectively. Having defined the detailed scores for each sub-route, we calculate the overall score $\left(O_{S,E}^{total}\right)$ for the proposed itinerary from point *S* (start) to point *E* (end).

$$O_{S,E}^{total} = \sum_{i,j=1}^{n} \left\{ \frac{O_{i,j}^{t} + O_{i,j}^{s} + O_{i,j}^{r}}{(a+b+c)*n} \right\} \quad (11)$$

For the calculation of $\left(O_{S,E}^{total}\right)$ we do not take into consideration the proposed cost of a transaction, due to the fact that the system evaluates it through its normalization. The evaluation of the cost is performed through the formula:

$$O_{i,j}^{cost} = \frac{cost_{i,j}}{min(cost_{i,j})} \quad (12)$$

where $min(cost_{i,j})$ represents the minimum cost for the specific route. At this point we encapsulate into the overall score the cost's score in order to recalculate a final score $\left(O_{i,j}^{final}\right)$ for the transaction, which will be the system's final recommendation to the user. It is:

$$O_{i,j}^{final} = \sum_{i,j=1}^{n} \left[ \frac{(O_{i,j}^{total} - O_{i,j}^{cost})^2}{f_{S,E}} \right] \quad (13)$$

### 4.3 An Example

This subsection presents an example of the recommendation process and its runtime environment. Having performed the optimal routes retrieval algorithm [4, 15], the user is transferred to the recommendation interface, where the results of the recommendation process are presented (Figure 6). At this phase, the evaluation of the itineraries is executed. More specifically, for every solution that has been retrieved for a requested transaction, the user may further consider its sub-routes. For each sub-route, the system calculates the average score that the carrier has received for its reliability during the transaction, as well as the average score for the transaction's duration. During the calculation of the above averages, the scores that each carrier (or each route) has received are multiplied by a user's reliability coefficient. This is performed in order to add a level of significance into a reliable user's opinion (compared with a less reliable one). Reliability refers to the number of times that a user has rated an itinerary, and not by the fact that his/her evaluation was considered as being strict or not. In addition to the above evaluation, a similar procedure takes place with respect to the safety and the overall carrier's reliability during the transaction. Both the average score of the specific elements (duration, reliability, safety, general reliability) and the overall score are stored in the system's database. When this procedure is completed for all itineraries' sub-routes, an average of all scores is extracted. The final score of the itinerary is the sum of the carriers' and the sub-routes' overall score, normalized by the overall cost and the number of intermediate transloadings. Moreover, the system retrieves information related to the completion of the above itineraries and their correspondent frequency. This procedure aims at checking whether a specific itinerary is constantly selected by other users. The popularity of each route is presented to the user later, in order not to affect his/her decision.

Initially, the recommended solutions are shown to the user according to their final score (top table of the interface shown in Figure 6). The user may then see each solution's details; by clicking on the "View Details" link (which appears at each entry of the top table), the interface expands dynamically and a second table appears (entitled "Sub-Route Details"), containing information about the sub-routes of the selected itinerary and the overall scores of each sub-route. Clicking on the "More Details" link, the user is provided with additional information about each sub-route (such as scores for its duration, safety and reliability). Moreover (by exploiting the "Show" link at the "Top-10 Carriers" column), the user is given the opportunity to compare a sub-route's carrier with any of the Top-10 carriers that exist for the particular sub-route (this is a common practice in CF techniques). In such a case, the interface of Figure 6 expands further and a third table, entitled "Top-10 Carriers", appears. When selecting a carrier from this table, by clicking on the "Select" link, the corresponding differences (in terms of cost, duration and carrier's rating) are presented in the bottom right part of the window (under the header "Additional Features").

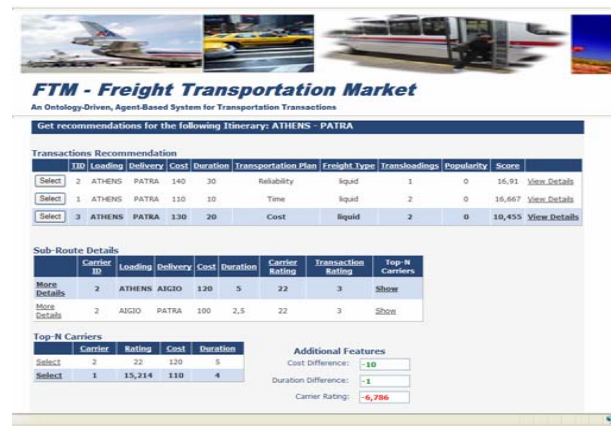

Figure 6: The recommendation module interface.

### 4.4 Implementation Issues

A new software agent, namely the *Recommender Agent (RA)*, has been implemented and interconnected with a correspondent Web Service, in order to coordinate the overall recommendation process. The main tasks of the RA concern the coordination of the recommendation module, depending on the characteristics of each transaction. Through these formally modeled tasks, RA provides continuous assistance to customers, while it remains active and capable to adapt its "behavior" into a rapidly changing environment. RA is responsible for the coordination of the whole process, as it interacts with the other software agents of the system [10]. Moreover, the recommendation policy of our system builds on Web Services concepts [26]. A Web Service is a URL-addressable software resource that performs functions and provides answers. It is constructed by taking a set of software functionality and wrapping it up so that the services it performs are visible and accessible to other software applications. A Web Service can be discovered and leveraged by other Web Services, applications, clients, or agents. In other words, Web Services can request services from other Web Services, and they can expect to receive the results or responses from those requests. Moreover, Web Services communicate using an easy-to-implement standard protocol (SOAP). Web Services may interoperate in a loosely-coupled manner; they can request services across Internet and wait for a response [5]. Due to the fact that external applications could exploit the proposed recommendation services, the implementation of the FTMarket's recommendation module was performed according to Web Services concepts and standards.

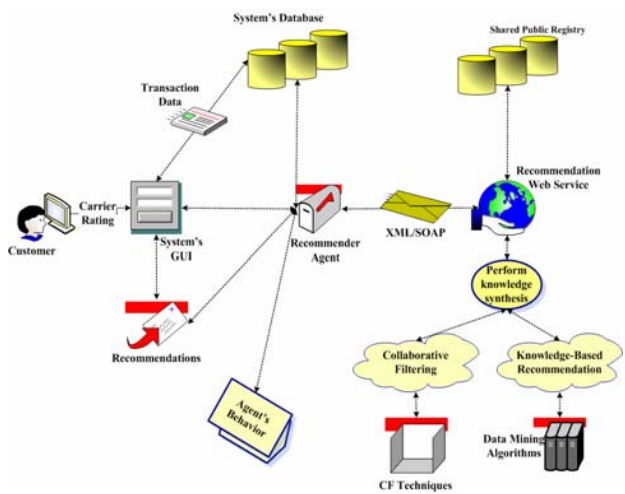

Figure 7: The recommendation module architecture.

The overall architecture of the FTMarket's recommendation module is illustrated in Figure 7. As shown, the module is appropriately wrapped in order to describe the kind of service to be provided. To be easily located by users, such descriptions of services are placed in a shared public registry. It is through this registry that users may look up for the services they need each time (in any case, a Web Service can be directly accessed if one knows its URL and WSDL). The correspondent agent that needs functions provided by the specific Web Service sends the appropriate request as an XML document in a SOAP envelope. This protocol can work across a variety of mechanisms, either asynchronously or synchronously. Web Services may make requests of multiple services in parallel and wait for their responses. The set of services to be provided in the FTMarket platform will be increased in the future (it will constitute a services repository). It is noted that it is not necessary for all these services to be provided through a single server; multiple servers, located in distinct providers, may be used. Finally, our system's Web Services are message-based. Interaction via message exchange means that instead of a client invoking functionality exposed as a Web Service, it sends a request to the Web Service to have the functionality invoked [7, 8]. In other words, what a Web Service exposes is the functionality of receiving a message. We have adopted a generic message interchange, which means that delivery of message content is independent of its format.

## 5. Conclusions

This paper has elaborated a series of issues related to the integration of hybrid recommendation techniques into an agent–based transportation transactions management platform. We proposed a hybrid recommendation module that combines different recommendation techniques in order to provide the user with more accurate and efficient suggestions. The overall recommendation process is coordinated by a software agent, which is responsible for carrying out multiple tasks, such as coordination of the recommendation module, selection of alternatives and knowledge synthesis through the exploitation of different recommendation techniques and algorithms. The presence of the *Recommender Agent* guarantees that the user will be provided with continuous recommendations, which are dynamically updated. Finally, we have exploited concepts related to Web Services in order to make the proposed recommendation functionalities accessible from external applications.

Future work plans mainly concern the consideration of additional recommendation techniques, such as content–based or model–based techniques and the exploitation of data mining algorithms in order to enhance the overall quality of the recommendations provided. The development of additional (local or remote) Web Services, which will be capable of carrying out more complex requests for recommendation techniques synthesis, is another major concern.

**Dr. Alexis Lazanas** studied Applied Informatics in Athens University of Economic and Business (B.Sc. 1996) and received his Ph.D. from University of Patras (Greece) in the field of Recommender Systems, Data Mining and Intermodal Transportation (2008). He worked in Technological Educational Institute (T.E.I.) of Patras as Scientific Collaborator and as Software Developer – Special Analyst in various major companies. Currently he is working as Teacher of Informatics in Greek Public Education. His research interests are on the areas of Agent-based Information Systems, Data Mining, Web Technologies, Hybrid Recommender Systems and Intermodal Transportation Management.